# Flame Stability Analysis of Flame Spray Pyrolysis by Artificial Intelligence

Jessica Pan[1]    Joseph A. Libera[2]    Noah H. Paulson[2]
Marius Stan[2]

[1]*Princeton University, Princeton, New Jersey, United States*
[2]*Argonne National Laboratory, Lemont, Illinois, United States*

October 20, 2020

**Contributor Email(s)**
Jessica Pan: jp41@princeton.edu (https://orcid.org/0000-0003-3724-4538)
Joseph Libera: jlibera@anl.gov (https://orcid.org/0000-0002-6226-3397)
Noah Paulson: npaulson@anl.gov (https://orcid.org/0000-0002-3548-9120)
Marius Stan: mstan@anl.gov (https://orcid.org/0000-0002-7087-7348)

# Flame Stability Analysis of Flame Spray Pyrolysis by Artificial Intelligence

Jessica Pan[1]     Joseph A. Libera[2]     Noah H. Paulson[2]

Marius Stan[2]

[1]*Princeton University, Princeton, New Jersey, United States*

[2]*Argonne National Laboratory, Lemont, Illinois, United States*

October 20, 2020

**Abstract.** Flame spray pyrolysis (FSP) is a process used to synthesize nanoparticles through the combustion of an atomized precursor solution; this process has applications in catalysts, battery materials, and pigments. Current limitations revolve around understanding how to consistently achieve a stable flame and the reliable production of nanoparticles. Machine learning and artificial intelligence algorithms that detect unstable flame conditions in real time may be a means of streamlining the synthesis process and improving FSP efficiency. In this study, the FSP flame stability is first quantified by analyzing the brightness of the flame's anchor point. This analysis is then used to label data for both unsupervised and supervised machine learning approaches. The unsupervised learning approach allows for autonomous labelling and classification of new data by representing data in a reduced dimensional space and identifying combinations of features that most effectively cluster it. The supervised learning approach, on the other hand, requires human labeling of training and test data, but is able to classify multiple objects of interest (such as the burner and pilot flames) within the video feed. The accuracy of each of these techniques is compared against the evaluations of human experts. Both the unsupervised and supervised approaches can track and classify FSP flame conditions in real time to alert users of unstable flame conditions. This research has the potential to autonomously track and manage flame spray pyrolysis as well as other flame technologies by monitoring and classifying the flame stability.

# 1 Introduction

Flame spray pyrolysis (FSP) is a cost-effective, versatile and scalable synthesis process used to generate powder-like nanoparticles by combusting a solvent loaded with precursors. In one widely used modality (which this study follows), the liquid precursor is atomized via oxygen flow and ignited by pilot flames [1]. The resulting nanoparticles may be oxides, salts, or metal/alloy nanoparticles. Notably, they can be synthesized at low cost with FSP and employed as reinforcing agents, pigments, catalysts, and battery materials [2]. Streamlining the FSP process by detecting and avoiding unstable flame states would result in less contamination (by soot or other waste products) and improve the purity of yield products.

For new FSP mechanisms involving novel chemistries or conditions, detailed investigation is needed to identify the particle formation mechanisms at play. In other words, the parameters required to produce or predict the properties of a target yield of nanoparticles are still unclear because of the lack of a complete understanding of the mechanisms by which they form [3]. Moreover, a contaminated product could be created if FSP precursors are not fully combusted due to unstable flame conditions and could become mixed with the collected material. Thus, it is advantageous to identify and avoid unstable flame conditions. Detecting and lowering the frequency of instability in FSP not only assures the production of better quality nanoparticles, but also contributes to more efficient workflows.

There have been multiple attempts to model FSP to better understand the important physical mechanisms of nanoparticle formation. A comprehensive analysis of the physics and chemistry of premixed flames includes the gas dynamics of curved flames and flame-induced flows as well as flame propagation in open or confined environments. The nonlinear stages of the hydrodynamic instabilities inherent to any premixed flame in a gaseous fuel are affected by compressibility, acoustic waves, and shock waves [4]. Early approaches using Brownian collision-coalescence models have been used to predict growth rates of particles in the spray [5]. In 2009, Widiyastuti *et al.* studied the formation of silica nanoparticles through solid-fed flame synthe-sis. The effects of processing parameters such as flame temperature, residence time, and precursor particle size were analyzed using both numerical modeling and simultaneous mass transfer modeling [6]. The study concluded that variables such as methane flow rate, carrier gas flow rate, and initial particle size all played a role in the efficacy of silica nanoparticle formation during FSP [6]. A review of recent trends in the improvement of flame aerosol synthesis by Li *et al.* discussed additional advancements in understanding the dynamics of gas-phase flame synthesis using in-situ optical dynamics and multi-scale modeling [7].

Flame stability was also examined using experimental monitoring and characterization tools involving digital imaging and computational spectral analysis [8]. The results demonstrated that flames are extremely unstable at premixed fuel-lean conditions. Further statistical analysis of flame images resulted in assessments of the stability of flame in terms of its color, geometry, and luminance [9]. Meanwhile, computational fluid dynamics (CFD) can provide insights into the combustion characteristics and flame stability at the microscale. For example, for the case of premixed methane/air mixtures, CFD simulations demonstrated that the main flame extinction modes include a spatially global type for large wall thermal conductivities and/or low flow velocities and blowout [10]. Even more recently, Dasgupta *et al.* used CFD modeling of nanoparticle synthesis in FSP [11]. This physics-driven approach has provided a keen understanding of the forces at play during the synthesis process by investigating the dynamics of the solvent, air, and flame around the FSP burner [11]. A similar study completed by Oliver-Martinez *et al.*, involving the synthesis of silica nanopowders used CFD to compute the magnitude and average particle diameter given different inputs (environmental parameters) and reactor sizes [12]. All of the methods outlined above employ physics and chemistry to analytically or computationally model the creation of nanoparticles during FSP and to determine their size and yield.

The present work did not examine in depth the physics behind FSP, but rather employed image processing techniques to analyze FSP footage. While modeling FSP flame and nanoparticle creation can help in the understanding the underlying physical systems and use such an understanding to predict FSP outcomes, in practice such methods would be too slow to provide a real-time estimate of FSP flame conditions. It would require the models in question to be re-run every time a parameter (such as oxygen flow rate, nozzle temperature, etc.) changes. Additionally, a given set of parameters does not completely describe the environment and the experimental setup. For example, changes in humidity or the quality of the precursors may cause flame stability to vary with the same nominal parameters. As a result, the state of the art for monitoring FSP flame conditions is direct human observation and supervision of the flame as FSP is in progress.

In this work, artificial intelligence elements such as machine learning and computer vision were used to devise a real-time method of detecting unstable flame conditions within seconds of their occurrence. In both unsupervised and supervised approaches, the algorithms were first trained and then used to generate predictions on unknown data within seconds. The unsupervised learning approach involved separating unknown, unlabelled data into clusters based on the variance between their respective features, and then labelling the data based on which clusters they were assigned to. This method did not require much human intervention because the training data was not prelabelled. Meanwhile, the supervised learning method required a large set of human-labelled data; in the case of this study, the data was then used to train an object detection model for classifying new images.

## 2  Data and Methodology

### 2.1  Flame Spray Pyrolysis

Experiments were performed using a FSP setup similar to that of Mädler *et al.* [1] at the Materials Engineering Research Facility at Argonne National Laboratory. A reagent alcohol solution ($C_2H_5OH$ + $CH_3OH$ 94.0 − 96.0%, $C_3H_8O$ 4.0 − 6.0%, VWR Chemicals BDH, USA) was sprayed from an oxygen atomizing nozzle (Schlick Nozzle Model 970 S8, D4.1016/1 Version 1.0, Germany). Combustion was initiated and sustained by oxygen/methane pilot flames. Additional oxygen was provided through a sheath flow. A Latin-Hypercube [13] design of experiments was employed to explore process variables critical to flame stability including the liquid flow rate, atomization $O_2$ flow rate, and sheath $O_2$ flow rate. This resulted in 53 successful experimental runs covering a large range of combustion regimes.

The image data was acquired using a Canon Rebel EOS XS digital camera (Figure 1) with a Semrock FF02-435/40-50 color line filter. The Semrock filter was chosen to highlight flame light emission centered on the CH* radical emission region as shown in Figure 2. By using the CH line filter, the image captured primarily the active reaction regions of the flame where the alcohol solvent was decomposing and releasing CH* radical species [14]. The spectrum of the line filter compared to the digital camera image collection and original light composition is shown in Figure 3. In the absence of a feature selective filter, the images only revealed silhouette images which are less informative of flame structure. The quantum efficiency of the CMOS sensor in the digital camera showed that the data was primarily collected in the blue channel of the RGB data. However, the data shown was the sum of all three RGB channels. A camera focus utility transferred images to the computer monitor at 25 hZ. Active Presenter software was used to record the data using screen capture video data acquisition.

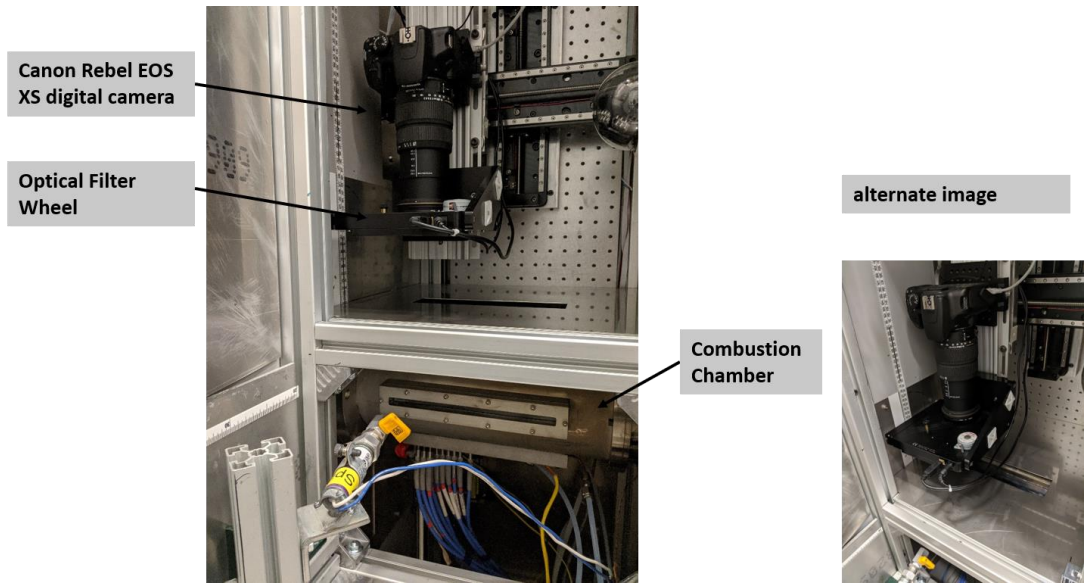

Fig. 1: *Setup of the Canon Rebel EOS digital camera in relation to the flame spray pyrolysis chamber*

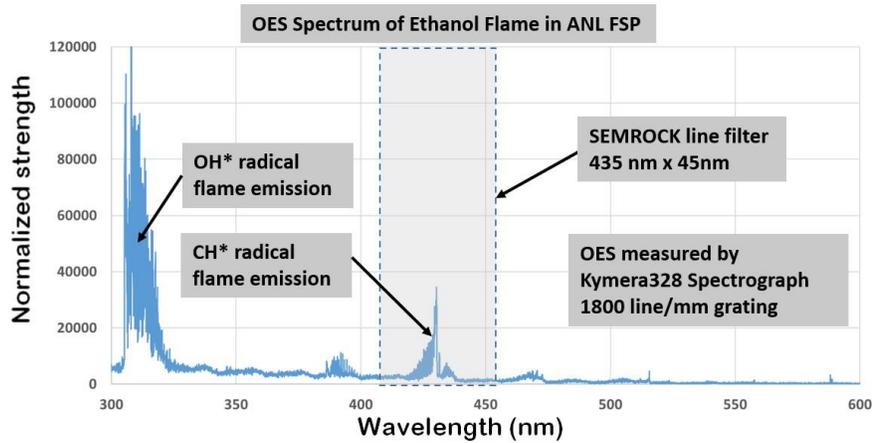

Fig. 2: *An optical emission spectrum as measured by an ANDOR Kymera 328 spectrograph using an 1800 line/mm grating*

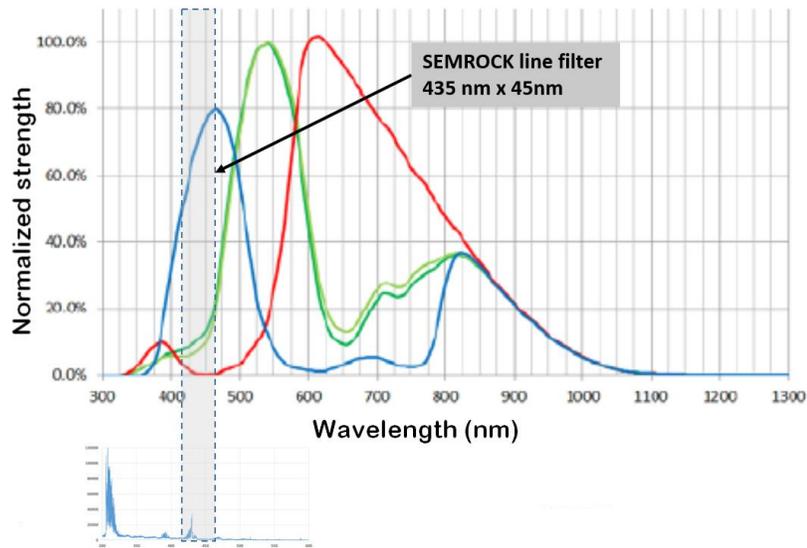

Fig. 3: *Red, green, blue (RGB) quantum efficiency of the Canon Rebel EOS CMOS sensor; the relationship between the digital camera image collection characteristics, light composition of the source flame, and SEMROCK line filter. The optical emission spectrum from Fig. 2 is shown below the quantum efficiency.*

## 2.2 Human Evaluation of Flame Stability

The increasing number of models that formally incorporate human expert opinion from a diverse range of experience, skills, and perspectives unavoidably led to a proliferation of informal use of expert opinion in ecology [15], forensic science [16], social sciences [17], and biology [18]. It is often accepted that an expert can be "anyone with either relevant and extensive or in-depth experience in relation to a topic of interest" [19]. An extensive literature review and classification of such approaches is beyond the scope of this work. It is important, though, to note that a broad definition of "human expert" and "expert opinion" is well accepted as part of current practice in various research fields. In physical sciences and engineering, the prevalent approach involves a set of ranking criteria, some with multiple levels, and a methodology for eliciting, aggregating and analyzing expert opinions [20]. In this work, we employed a methodology consisting of the following steps:

*Measure selection* consisted of identifying image characteristics that are relevant to the process, are easy to detect by humans and can improve the reliability and predictability of the approach. Since one of the objectives of the study was to establish ground truth around the subjective human perception of flame stability, the measures were centered on the following main characteristics: flame shape, intensity, flame front and anchor point (with other considerations left at the discretion of the experts). The boundaries between stable, unstable, and uncertain were left at the discretion of the experts, with no quantitative evaluation required.

*Selection of human experts* was aimed at providing a broad set of perspectives and evaluation styles. The evaluation group consisted of 11 Argonne researchers, Princeton professors, and students, all of whom are familiar with combustion and able to identify the stability of flames. The team included: two Argonne researchers who have direct experience with FSP; two Princeton professors with expertise in combustion and thermodynamics; two Princeton graduate students with expertise in combustion and thermodynamics; and four Princeton undergraduate students currently studying mechanical and aerospace engineering and who have taken courses in combustion and thermodynamics. To preserve the independence of evaluations, the authors of this study did not participate as human experts.

*Expert opinion elicitation* involved a questionnaire that required each of the experts to analyze a set of 53 video files of FSP flames. They were then asked to classify them in a database by assigning '2' for a stable flame, '1' if they were uncertain, and '0' for an unstable flame. Participants did not collaborate or discuss their evaluations with others. In discussions that followed the elicitation, a large majority of experts described watching fluctuations in either the flame front or anchor point as their main method of determining the stability of the flame.

*Expert opinion aggregation* consisted of analysis of data collected through the expert opinion elicitation. The human expert ratings were averaged across each video and the resulting distribution was treated as ground truth. This analysis is described in detail in Section 4.

## 2.3   Fluctuating Luminance Stability Classifier

A computational approach to flame stability classification was necessary for this project to generate predictions that require minimal human interference. Although human evaluations of these clips were available, an objective method to establish the flame state in different video clips provides an alternate approach. The flame state of new video clips was quantitatively established using this method. The area closest to the nozzle, where the flame originates, is called the anchor point. Momentary extinction events occurring at the anchor point are indicators of flame instability. In the footage, extinction events were indicated when the flame front separated from the nozzle (see Figure 4). Using this knowledge, it was possible to develop a protocol to quantitatively define flame stability of new footage after analyzing the anchor point regions for extinction events.

The devised method for quantitatively determining flame stabilization (referred to as the fluctuating luminance stability classifier, or FLSC) involved analyzing the luminance (a measure of pixel brightness) of pixels within a predefined bounding box region around the anchor point of the burner flame such as in Figure 4. A pixel's luminance was stored as an integer value between 0 and 255, with 0 indicating the darkest possible pixel and 255 being the brightest. The bounding box used in this study was 50 pixels tall and 30 pixel wide; it was centered around the nozzle with the top-left corner 450 pixels right from the leftmost border and 270 pixels from the bottom-most border, such that it fully encased the flame's anchor point. The area closest to the nozzle exhibited fluctuations in brightness when the flame was not stabilized due to the occurrence of momentary extinction events where the flame was no longer anchored to the nozzle. With this in mind, the live footage was processed to detect changes in luminance and used to classify the flame stability.

First, the luminance of all of the pixels in a bounding box across all of the frames in a given video clip was averaged to generate a mean bounding-box-pixel luminance value across the entire clip. Each frame's deviation in luminance from the calculated mean in the video was then computed. If the fluctuation in luminance deviated more than 25% from the average, the flame in the clip was considered unstable. If it deviated more than 15% from the average but less than or equal to 25%, the flame was considered to be of 'uncertain' stability. Otherwise, the flame was considered stable. This process as applied to example videos is shown in Figure 5b. The threshold percentages were chosen using visual inspection of the video clips in the dataset and their resulting FLSC plots.

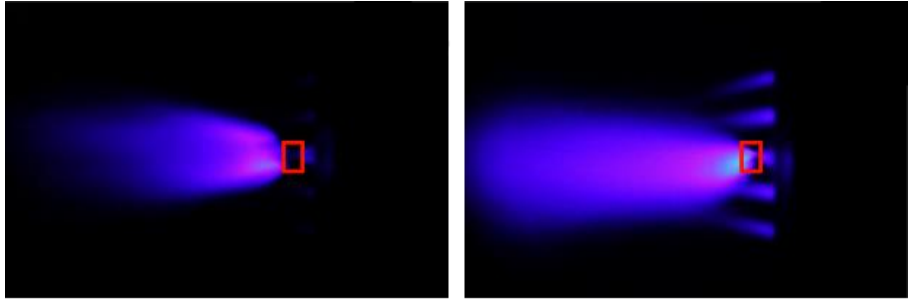

Fig. 4: *Unstabilized flame (left) and stabilized flame (right). The un-stabilized flame front is no longer attached to its origin (the nozzle towards the right of the image), while the stabilized flame front is still anchored to its nozzle. The approximate position and size of the bounding box used in this study is outlined in red.*

This thresholding method was effective at characterizing flame stability from video data because the mean bounding box luminance across a given video clip was ascertained and the luminance of individual frames was then compared to the overall mean. Suppose the flame was anchored to the nozzle for the vast majority of the clip – the mean bounding box luminance would therefore be relatively high as the pixels in the bounding box would be illuminated due to the brightness of the flame at the anchor point. If an extinction event occurred, the pixels in those respective frames would have low luminance. After comparing this low luminance to the overall luminance of the clip, it would be very likely that these frames would cross one of the preset thresholds, and the clip would be (correctly) labelled

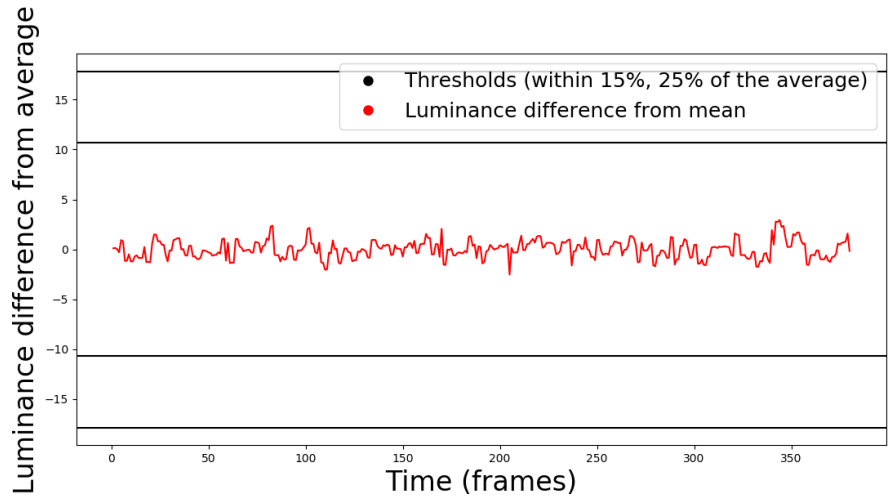

(a)

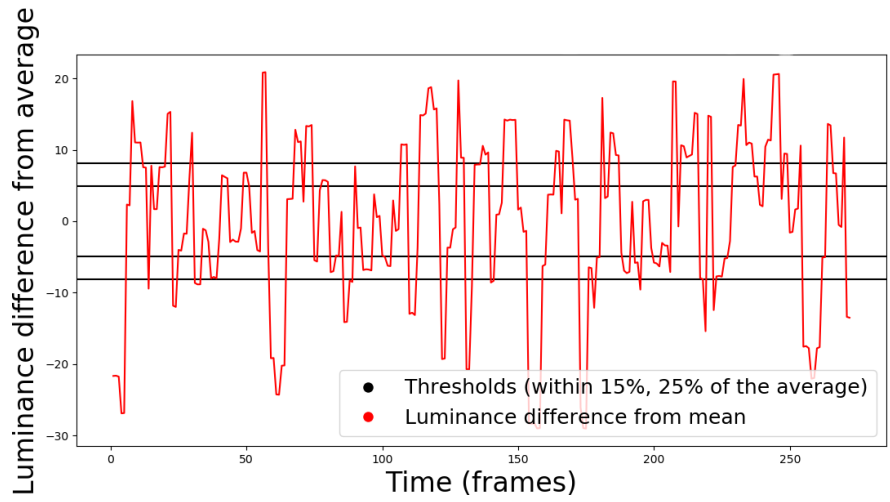

(b)

Fig.5: *Bounding box luminance fluctuation plot for an a) stable flame, b) unstable flame.*

'unstable'. Conversely, if a flame was not anchored to the burner nozzle for much of a given clip (which would still be indicative of an unstable flame), the value of the mean bounding box luminance associated with the sample would be low. If an anomalously bright pixel reading registered within the bounding box due to sensor noise, the given clip might be erroneously labeled as 'unstable' due to this large deviation from the mean. When evaluating an entirely still portion of the video clip (an area without the flame), sensor noise accounted for a maximum luminance deviation of 0.4 from the average luminance across the entire clip. This makes erroneous detection of an extinction event due to sensor noise extremely unlikely.

## 2.4 Unsupervised Machine Learning

The FLSC method predicts the stability of the flame by averaging pixel luminance data across a video clip and comparing the average luminance of each singular frame to the global average. The only way to obtain this global average is to first run and process the baseline video footage. For this to be done within a reasonable time frame (within seconds), the global clip would need to be very short and thus have very few (30-60) frames; such a clip would not long enough to establish a baseline and reliably catch outlier frames. There must exist a way to classify new data without needing to process the entire sample first. An unsupervised machine learning approach can be used to find patterns in an unlabelled dataset without human input. In this work, the features of interest (which will be described in more detail later) were composed of bounding box pixel luminosities in consecutive frames of the video feed so as to include features representative of the flame dynamics.

Principal component analysis (PCA) is a common dimensionality reduction technique that enables the compression of data from an $n \cdot p$-dimensional space (where $n$ and $p$ are positive integers) into a much smaller one while minimizing information loss [22]. Specifically, PCA finds linear combinations of the input dimensions such that the first combination, or principal component, maximizes the variance in the data, the second principal component maximizes the variance on an axis orthogonal to the first principle component, and so on with each additional principal component. Consequently, this principal component representation can be truncated with minimal information loss. In this study, each unsupervised learning feature was composed of a vector containing the luminance of every pixel in the bounding box region of a one-second-long video clip (corresponding to 30 consecutive frames of footage). So, the original feature space of each of the data points was 45000-dimensional, which was then reduced to merely 2 principal components using PCA. As a result, utilizing PCA in this study aided in the visualization of data and enabled the discovery of groupings or patterns within data that may have otherwise eluded notice due to the high dimensionality.

Additionally, k-means clustering is often used to group unlabelled data samples into $k$ clusters (where $k$ is a predefined positive integer) based on their shared features. This study followed Lloyd's algorithm for k-means clustering [24]. Lloyd's algorithm initializes $k$ centroids randomly in the feature space. The location of each data point on the plane is then compared to the location of each of the centroids – the point is then grouped with the centroid that it lies closest to. After each point is assigned, the location of the initial centroid is updated so that it becomes the centroid of all of the points assigned to it. The algorithm terminates when all of the points have been assigned a centroid and the location of the centroid no longer changes during the re-calculation step.

In this study's unsupervised learning approach, the plane formed by the 2 principal components formed the space upon which $k = 3$ random centroids were initialized. $k = 3$ clusters were chosen because the data is expected to fall into one of the three categories 'unstable', 'uncertain' or 'stable'. Once the k-means algorithm terminates and all points have been assigned to a cluster, any potential trends or patterns in the data points can be determined by assigning labels (of 'unstable', 'uncertain', or 'stable') to each of the points and deciding whether similarly-labelled points fall into the same clusters. Python's scikit-learn, scipy, and opencv libraries were used extensively to process and classify images for FLSC and the unsupervised learning approach.

## 2.5 Supervised Machine Learning

Object detection and image classification are widespread techniques used in artificial intelligence and computer vision. They can be used to quickly identify features or objects captured in either a still image or in a video feed. In this work, such a model was trained to classify stable and unstable flame conditions in FSP footage. These results may align more closely with human classifications than the unsupervised learning model as it was trained with images labelled according to the authors' own visual (and human) evaluation of each frame. FLSC could not be used to classify the images used for this supervised learning method because it cannot define all of the features of interest within each image, nor can its input be still images.

TensorFlow [25] (in particular, TensorFlow's Object Detection API and model zoo) was used to train the region-based convolutional neural network employed in this study to detect and classify images in video feed. The software employed in this component of the study is a modification of a tutorial developed by Jura et al. [26].

TensorFlow's RCNN Inception V2 model was used to extract and label different features in the video data [25]. The object detection process proceeds in two steps, a) detection of regions/objects in the original image, and b) classification of the regions/objects. The model first receives an input image and runs it through a neural network. Convolutional layers with pretrained filters process each image, generating a region-independent feature map. This feature map is then used as input to a region-wise multi-layer perceptron [27]. For the purpose of this study, the objects of interest were the 'burner cap', 'pilot flame', 'unstable [flame]', and 'stable [flame]'. The Softmax activation function was the final layer of the neural network and performed the classification [28]. The Softmax layer's inputs were the real values generated by the last CNN layer which were transformed into a distribution of probabilities (each probability being the chance the image represented one of the previously enumerated objects) [29].

One in every 30 video frames (images) of the entire dataset were randomly sampled and each of the relevant features (stable flame, unstable flame, pilot flames, burner cap) were manually labelled to compile the training and test sets needed to train the model used in this supervised learning approach. This resulted in a set of approximately 570 images, 20% of which were allotted to the test set and the rest of which formed the training set and were used to modify the weights in the underlying Softmax layer that classified features of interest. The RCNN Inception V2 model was trained using this labelled set of images for approximately thirteen hours; afterwards, it was run on unseen footage so it could generate predictions on the new data.

## 3    Results and Discussion

This section discusses the results of both the unsupervised and supervised machine learning algorithms. In the unsupervised approach, the original 45,000-dimensional feature space was reduced to just 2 principal components while still capturing a cumulative 70% (58% in principal component 1 and 12% in principal component 2) of variance in the data. These points were then grouped into three clusters using $k$-means clustering.

There was a correlation between a video clip's principal components and its predicted stability. These trends were visualized by coloring the data points according to their assigned classifications (stable, unstable, or uncertain) using FLSC and human evaluations as seen in Figure 6a and 6b, respectively. The originally labelled data was split into 3 categories (stable, unstable, uncertain), and so the points were grouped into 3 clusters using k-means clustering to separate the 'unstable' cluster from the other (unlabelled) clusters.

The stability of new data was ascertained by determining where they lie in relation to known stable/unstable/uncertain clusters in principal component space. If a point was closer to the centroid of the 'unstable' region than the other two centroids, then it was labeled 'unstable'. Otherwise, it was labeled 'stable'. Thus, to label new data, the algorithm required 30-frame video clips of FSP to transform into the predetermined principal component space. The program referenced the k-means centroids to assign the data point to a cluster.

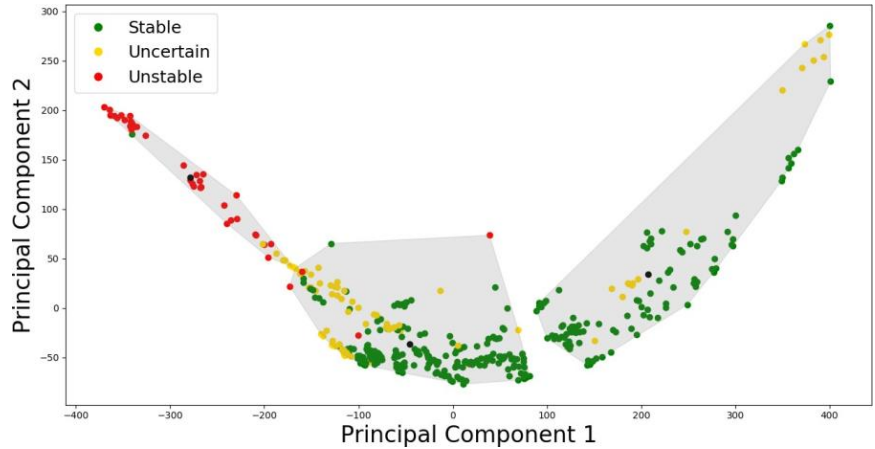

(a)

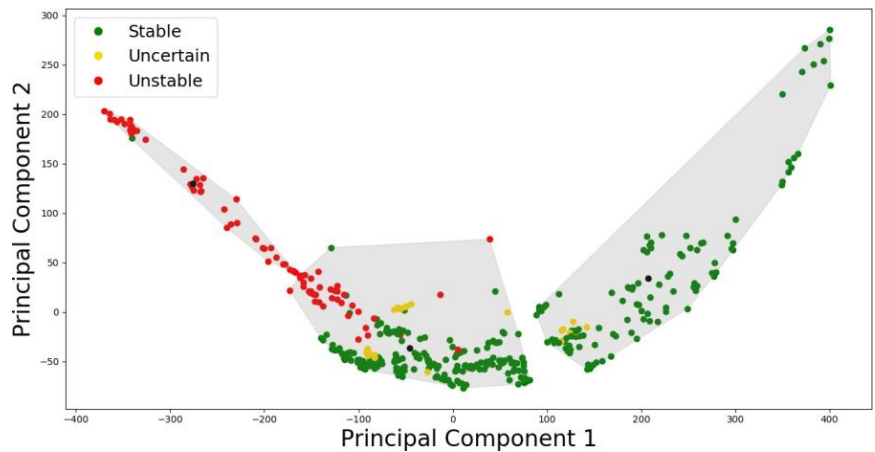

(b)

Fig. 6: *PCA results for bounding box pixel luminance are plotted with each point representing 30 consecutive frames in a given video clip, or one second of footage, and colored according to the a) FLSC, and b) human evaluation approaches. Black points represent the centroids of the three main clusters which are wrapped in gray convex hulls. The leftmost cluster contains nearly all of the unstable data and so it will be considered the 'unstable' cluster. The other two clusters are not labelled.*

Unlike unsupervised learning, the supervised learning process required much more human intervention, particularly with the task of manually labelling training data. Approximately 3.3% of all of the frames in the dataset were randomly sampled to create the test and training sets for training the RCNN Inception V2 object detection model. Afterwards, the relevant features in the image (stable flame, unstable flame, pilot flames, burner cap) were bounded and labelled before the RCNN V2 model was trained on the labelled images. Afterwards, the model was given new data so it could detect and classify the features of interest in that image (such as in Figure 7).

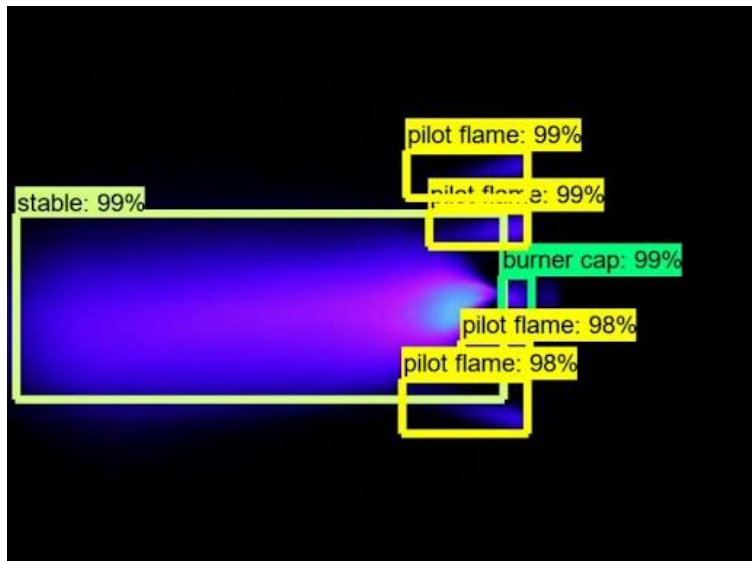

Fig.7: *Image with features detected and classified by trained object detection model.*

In Figure 7, the supervised learning model was able to recognize and label features such as flame state, pilot flames and burner caps in new images; the numbers associated with the labels in each boxed region represent the percentage match of the features detected with their respective classifications. The user may use these to define min-imum thresholds for a match with an 'unstable' classification that, if exceeded, could warn users about unstable flame conditions.

The predictions generated by the fluctuating luminance stability classifier and both the unsupervised and supervised learning approaches are compared with human evaluations to measure the success of these methods. The numerical human classifications for each video clip were averaged for the 11 subjects and the resulting mean was recorded. To directly compare the human classifications with the machine-generated predictions, any averaged human classification greater than 1.2 (more than 'uncertain' stability classification) was considered 'stable', and any averaged classification less than 0.8 was 'unstable' (if the average was between 0.8 and 1.2, the human classification was given an 'uncertain' value of 1). This numeric labelling was extended to both the fluctuating luminance stability classifier, unsupervised learning, and supervised learning approaches. Note that the unsupervised learning method classified footage one second at a time; if the prediction of any one-second fragment belonging to a video clip was classified 'unstable', then the prediction for the entire clip would be considered 'unstable' (it was labelled 'stable' otherwise). Figure 8 is a compiled visualization of the resulting predictions for the 53 videos in the dataset using FLSC, unsupervised learning, and supervised learning approaches compared to human evaluation.

The range of predictions was represented on a continuous scale from 0 to 2. To develop statistics of the accuracy of each approach as compared to the ground truth every prediction greater than 1.2 was considered stable and every prediction less than or equal to 1.2 was considered unstable (this study treated all 'uncertain' predictions as potentially unstable flame states). FLSC matched human predictions for 46 of the 53 videos, or 86.8% of the time. The prediction achieved using the unsupervised machine learning model matched that of the human classifications for about 73.6% of the cases. The accuracy of the supervised machine learning model was even higher, with human experts and the model agreeing in about 90.6% of the clips. As a baseline, 1000 random trials in which each video was uniformly randomly assigned a 'stable' or 'unstable' classification, resulted in an average of 49.8±6.9% of the videos being labelled correctly.

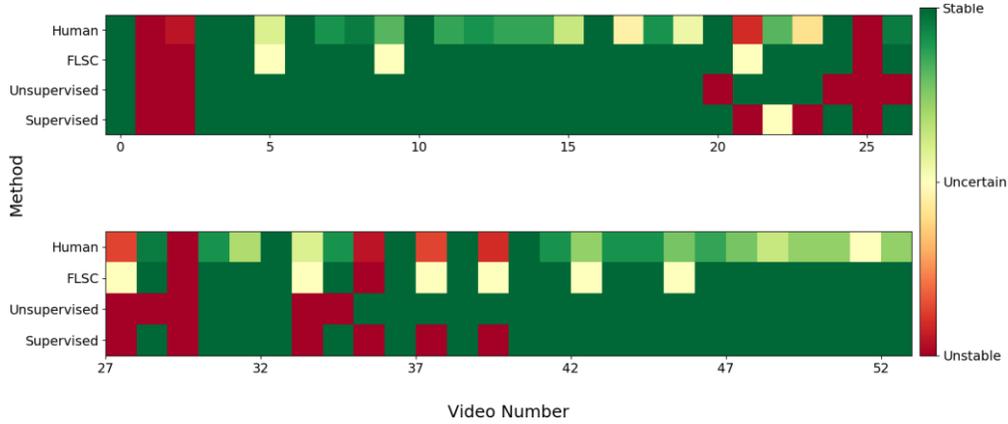

Fig. 8: *Comparing stability classifications of human experts to that of the fluctuating luminance stability classifier (FLSC), unsupervised learning and supervised learning predictions.*

Both supervised and unsupervised machine learning techniques have their limitations and drawbacks. The unsupervised machine learning model is not easily generalizable to different types of video feed. If the angle, shape, or other aspect of the video itself is changed, then humans are required to re-define another bounding box around the nozzle manually in order to run the analysis again. This is because the bounding box luminance stability classification is dependent on the luminance of pixels near the burner nozzle, which is not automatically defined by the program and requires human manipulation.

Generalizability is a lesser concern for the supervised image detection and classification approach – the computer vision algorithm should be able to classify flame stability despite some changes in the way the FSP video is recorded (angle, size, etc). This is because RCNNs analyze patterns in regions of each frame to classify images rather than focusing on a specific part of the image itself. The model was only trained on video clips taken with the one setup described in the methodology section. As a result, the accuracy of the model's predictions will likely decrease on videos of different proportions or angles if it does not undergo further training on a more diverse set of data. For example when the algorithm was tested on video clips of FSP running at a different angle and under different lighting conditions than the one used in this study, and the model indeed misclassified or failed to detect elements in those clips.

It may be possible that the predictions generated by machine learning models are more accurate than human evaluations in some cases, while in other cases they fail to detect unstable flame states. The false positive rates (where a positive outcome was 'unstable') for the FLSC, unsupervised learning, and supervised learning approaches were 7.9%, 13.2%, and 2.6%, while the false negative rates were 26.7%, 60%, and 26.6%, respectively. Cases where humans gave a stable evaluation to a video clip while FLSC and the unsupervised and supervised learning models considered the same clip unstable may be attributed to the human participants' inability to detect an extinction event that passes too quickly. Video clips were analyzed frame-by-frame by the program, and therefore the algorithm detected extinction events on a very small time scale. Meanwhile, the high false negative rates for the FLSC and unsupervised learning predictions may be due to the arbitrary thresholds that were set to quantitatively determine the stability of the video clips. Changing the thresholds to find the optimal classification may improve FLSC as well as unsupervised learning predictions.

Another potential cause for the higher false negative rates may lie in the data itself. The size of the dataset may be too small to achieve a full feature map for the unsupervised learning algorithm to create informative clusters with which to generate its predictions. Also, as a result of using the prediction thresholding approach mentioned earlier (where each prediction greater than 1.2 is considered stable and each prediction less than or equal to 1.2 is considered unstable), 15 of the 53 video clips are labelled 'unstable' by the ground truth while the remaining 38 are considered stable (more than twice that of the unstable data set). This bias in the dataset may also skew the predictions of the object detection algorithm used in the supervised learning approach to favor stable predictions.

# 4  Conclusion

This study's objective was to detect unstable flame states which could result in sub-optimal or impure yields of target particles during FSP. Real-time optimization of manufacturing methods can boost productivity and enhance workflow by utilizing artificial intelligence to monitor its processes. Stability monitoring approaches such as those developed in this work play a critical role in controlling flame stability for FSP with the potential to increase the quality and quantity of nanoparticles produced with benefits to energy, security, and health industries.

A fluctuating luminance stability classifier was developed to classify the stability of short video segments of the flame through detecting variations in brightness near the burner flame nozzle. An unsupervised learning algorithm based in PCA and k-means clustering on FSP video feed data was developed to categorize flame stability without the need for human labelling of training data. Additionally, to achieve better accuracy for the labelling of specific FSP components, an object detection classifier was trained to recognize and identify features in videos of FSP. This supervised learning approach required prior human labelling of test, validation, and training data on which to train the object detection classifier.

Overall, the fluctuating luminance stability classifier, unsupervised, and supervised learning predictions achieved 86.8%, 73.6%, and 90.6% accuracy, respectively, when compared to human evaluations of the same data (predictions assigned randomly were correct approximately 49.8% of the time). While human error and differences in opinion may account for some of the resulting inaccuracies, expert classifications should still be considered ground truth because traditionally humans monitor FSP and evaluate its flame condition with good success rates. However, it should be noted that sophisticated computational approaches may be able to accurately detect instability events that the human eye cannot. Continuation of this research may include developing an autonomous approach to control the FSP input parameters to maintain flame stablity (rather than simply alerting researchers if the flame is unstable), or using a different object detection model such as YOLO (You Only Look Once), which has been shown to outperform Fast/Faster RCNN like the one used in this study [30]. This study shows that visual features of the FSP flame combined with machine learning and computer vision may contribute towards a real-time, artificial-intelligence-powered model which improves the efficiency and efficacy of flame spray pyrolysis.

# 5   Declarations

## 5.1   Acknowledgements

The authors acknowledge Debolina Dasgupta and Jia Deng for useful discussions and guidance.

## 5.2   Funding

The authors acknowledge financial support from Laboratory Directed Research and Development (LDRD) funding from Argonne National Laboratory, provided by the Director, Office of Science, of the U.S. Department of Energy under Contract No. DE-AC02-06CH11357.

## 5.3   Conflicts of Interest / Competing Interests

The authors have no conflicts of interest to declare.

## 5.4   Availability of Data and Material

The data and material used in this study will be available upon acceptance of this manuscript.

## 5.5   Code Availability

The code used in this study will be available upon acceptance of this manuscript.